# DEEP CONVOLUTIONAL NEURAL NETWORKS FOR IMAGING DATA BASED SURVIVAL ANALYSIS OF RECTAL CANCER

*Hongming Li[1], Pamela Boimel[2], James Janopaul-Naylor[2], Haoyu Zhong[2], Ying Xiao[2], Edgar Ben-Josef[2], and Yong Fan[1]*

Departments of Radiology[1] and Radiation Oncology[2], Perelman School of Medicine, University of Pennsylvania, Philadelphia, PA, 19104, USA

## ABSTRACT

Recent radiomic studies have witnessed promising performance of deep learning techniques in learning radiomic features and fusing multimodal imaging data. Most existing deep learning based radiomic studies build predictive models in a setting of pattern classification, not appropriate for survival analysis studies where some data samples have incomplete observations. To improve existing survival analysis techniques whose performance is hinged on imaging features, we propose a deep learning method to build survival regression models by optimizing imaging features with deep convolutional neural networks (CNNs) in a proportional hazards model. To make the CNNs applicable to tumors with varied sizes, a spatial pyramid pooling strategy is adopted. Our method has been validated based on a simulated imaging dataset and a FDG-PET/CT dataset of rectal cancer patients treated for locally advanced rectal cancer. Compared with survival prediction models built upon hand-crafted radiomic features using Cox proportional hazards model and random survival forests, our method achieved competitive prediction performance.

*Index Terms*— CNNs, proportional hazards model, survival analysis, rectal cancer, tumor recurrence

## 1. INTRODUCTION

Colorectal cancer is the 3rd most common cause of cancer death in US. The standard of care for locally advanced rectal cancer is neoadjuvant chemoradiation therapy (CRT) followed by total mesorectal excision. The tumor response to CRT is heterogeneous with approximately 15-20% achieving a pathologic complete response (pCR) and most others achieving various degrees of partial response [1]. Patients who achieve a pCR have very favorable outcomes while those without pCR often develop local recurrences and distant metastasis. Since high-risk patients would be candidates for more intense or different systemic therapies, it would be very useful to be able to predict the individual risk of each patient. However, current predictive models based on clinical factors such as tumor stage, nodal stage, and hemoglobin do not have sufficient fidelity [2].

Radiomics is a promising tool for early selection of optimal treatment for patients. Recent radiomic studies in rectal cancer have demonstrated promising performance for cancer staging and treatment outcome prediction [3, 4]. Most of the existing studies build prediction models on hand-crafted radiomic features, such as tumor intensity histogram, shape, and texture patterns [5, 6]. In these studies, a large number of imaging features are often extracted, feature selection and dimensionality reduction techniques have to be adopted to relieve curse of dimensionality. A special effort is also needed to fuse multimodal imaging features [7]. Recent radiomic studies have witnessed promising performance of deep learning techniques in learning radiomic features and fusing multimodal imaging data [8-10]. In these studies, convolutional neural networks (CNNs) are widely adopted to learn informative imaging features [11]. However, it is not straightforward to apply the prevalent CNNs to tumor images since the CNNs require a fixed image size in both training and testing while the tumor size of different patients varies greatly. Although resampling different tumors to have the same size is a simple solution, elegant tools such as a spatial pyramid pooling strategy might lead better performance [12].

Most existing deep learning based radiomic studies build predictive models in a setting of pattern classification [8-10]. Although the pattern classification setting is a default for diagnosis, it might not be appropriate for survival analysis studies where some data samples have incomplete observations. It is common in cancer studies that many patients do not have complete observations, and data censoring has to be applied to such data samples. For such datasets, Cox proportional hazards model (CPH) is a popular model for survival analysis [13]. The CPH model is built with a linear assumption that a patient's risk of an event occurring is a linear combination of observed measures. Random survival forests (RSF) is another popular method for survival analysis [14]. Different from CPH, RSF is virtually free of model assumptions and capable of capturing non-linear effects or higher order interactions between the risk to be predicted and the available measures. Since both the methods build predictive models upon given features, their performance is hinged on the features to be used. Deep learning based survival analysis tools have also been developed and achieved better prediction performance than CPH and RSF models [15, 16]. Recent studies have also proposed CNNs based tools for pathological image based survival analysis, and promising performance have been obtained [17]. However, they are designed to handle image with fixed image size.

In this study, we develop a deep learning method under a flexible framework to learn discriminative radiomic features from multi-modal imaging data in a proportional hazards model setting. Our method is built upon deep CNNs to learn imaging features by optimizing partial likelihood of a proportional hazards model for survival information. The learning of imaging features based on CNNs is capable of capturing complex relations between imaging data and survival information, therefore may achieve better prediction performance than models built upon hand-crafted imaging features. Particularly, in our deep learning model multimodal 3D imaging data of tumors are fused by 3D convolutional layers in a data-driven way during the training of the model. Tumor size differences across subjects are handled by a spatial pyramid pooling layer [12]. Moreover, biological/clinical measures if available can also be seamlessly fused in the model by fully connected layers.

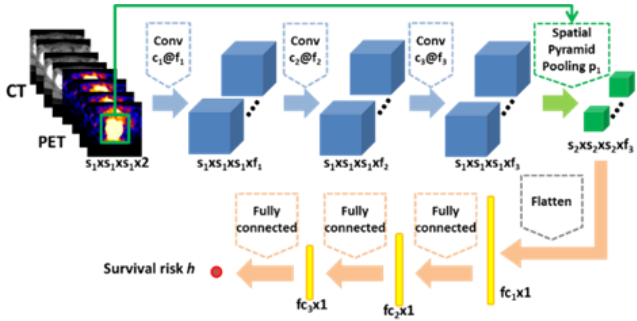

Fig. 1. The CNN architecture for survival risk prediction from multi-modal imaging data.

## 2. METHODS

Our deep learning based survival model is tailored for radiomic survival analysis based on the proportional hazards model and deep CNNs, which is expected to capture complex relationship between imaging information and survival information effectively.

### 2.1. Proportional hazards model

Given a survival dataset $D$ containing $N$ data samples, each sample $s_i$ ($1 \leq i \leq N$) has its feature vector $x_i \in R^M$, observed time $T_i$ for the event of interest, and event indicator $E_i$ that indicates whether the event occurred or not (right-censored). Under the proportional hazards assumption [13], the hazard function for $s_i$ is formulated as

$$\lambda(t|x_i) = \lambda_0(t)\exp(h(x_i)), \quad (1)$$

where $\lambda_0(t)$ is the baseline hazard function, and $h(x_i)$ is the estimated risk of $s_i$. The hazard function $\lambda(t|x_i)$ measures the hazard rate for sample $s_i$ at time $t$. The partial likelihood of all the observed (non-censoring, $E_i = 1$) events that have occurred to the corresponding samples is calculated as

$$L_D(h(x)) = \prod_{i:E_i=1} \frac{\exp(h(x_i))}{\sum_{j:T_j>T_i} \exp(h(x_j))}, \quad (2)$$

where $h(x)$ is the underlying risk function to be determined. To determine the $h(x)$ that best characterizes relations between the features and survival risks, maximum partial likelihood is adopted to optimize the model parameters in $h(x)$. The standard CPH model adopts $h(x) = \beta^T x$ which assumes a linear relation between the features and risks, and $\beta$ is optimized to build the predictive model. However, the linear assumption might not be able to capture complex relations between tumor images and clinical outcomes. To extract informative imaging features to predict the survival risk, we adopt deep CNNs to learn imaging features $x$ and $h(x)$ in a data-driven way.

### 2.2. Deep CNNs for predicting survival

To build a survival prediction model upon imaging data, we adopt deep CNNs to extract informative features that optimize partial likelihood of the proportional hazards model. Our model mainly contains 3 types of layers, including convolutional layers to extract informative imaging features, a spatial pyramid pooling (SPP) layer [12] to handle tumor size differences across patients, and fully connected layers to learn a nonlinear mapping between imaging data and survival risk. As illustrated by Fig. 1, our model contains 3 convolutional layers, followed by 1 SPP layer, 2 fully connected layers, and an output layer. The output layer contains one node to calculate the survival risk $h(x)$ for the input imaging data. Rectified linear units (ReLu) are used as the nonlinear activation functions for the convolutional and fully connected layers. Dropout is applied to the output of the fully connected layers [11]. The imaging scans for each patient are used as input to the deep learning model, and a bounding box containing the tumor region will be used as auxiliary input to the SPP layer. The SPP layer was implemented as a one-layer max-pooling with pre-set output size. Negative log partial likelihood of Eq. (2) is used as the loss function to train the whole network

$$\text{loss}(h(x)) = -\sum_{i:E_i=1}(h(x_i) - \log \sum_{j:T_j>T_i} \exp(h(x_j))). \quad (3)$$

The whole network is trained using stochastic gradient descent [11]. To train the survival prediction model for applications with limited training data, data augmentation is adopted to boost its robustness and performance. Particularly, augmented tumor imaging data are generated using image translation, rotation, and zooming [11]. The survival information for the augmented tumor images are sampled from a Gaussian distribution with the survival information of the original image as its mean value and a pre-set standard deviation.

## 3. RESULTS

We first validated our method based on simulated 2D imaging data with survival data generated with a nonlinear survival risk function of imaging information to demonstrate the performance of the proposed deep CNNs based survival model. Then, we applied our method to a PET/CT dataset of rectal cancer patients to validate its performance in real clinical applications based on multi-modal imaging data. Our method was implemented based on Keras [18].

Our network contained 3 convolutional layers, followed by 1 SPP layer, 2 fully connected layers, and 1 output layer. In particular, the numbers of filters were 16, 36, and 64 for each convolutional layer, and the kernel sizes were $3 \times 3$, $5 \times 5$, $5 \times 5$ for the simulated 2D images and $3 \times 3 \times 3$, $5 \times 5 \times 5$, $5 \times 5 \times 5$ for the 3D PET/CT scans. For the SPP layer, the minimum bounding box containing the region of interest was used as auxiliary input ("digits" in the simulated images, and tumor regions in the PET/CT scans), the bounding boxes have varied sizes across images, and the output sizes of the SPP layer were $8 \times 8$ for the simulated images and $8 \times 8 \times 8$ for the PET/CT scans. The numbers of nodes in the fully connected layers were 500 and 100 for the simulated dataset, 1000 and 200 for the PET/CT dataset.

We compared our method with both CPH and RSF methods. Hand-crafted radiomic features were used as the input to these two methods. We computed both non-texture features and texture features using a radiomic method, capable of extracting multimodal imaging features [7]. Different extraction parameters including wavelet band-pass filtering and quantization of gray levels were tested. The dimensionality of radiomic features was reduced using principal component analysis (PCA) for the CPH model to improve its robustness on the PET/CT dataset due to its relatively small sample size, and different dimensions including 10, 20, 40, 60, 80 were tested to obtain the best performance. Different numbers of trees including 50, 100, 200 and minimum terminal node size including 3, 5, 10 were tested for the RSF model and the best result was reported. Concordance-index (c-index) [19] was used to evaluate the prediction performance obtained by different methods under the same cross-validation setting.

### 3.1. Experiment on a simulated dataset

The simulated dataset was generated as illustrated by Fig. 2. In particular, we randomly selected 5000 images of "0" and 5000 images of "6" from MNIST handwritten digit database. Then, a random tumor weights $M \in R^{W \times H}$ was generated and multiplied with each of those digit images $I \in R^{W \times H}$ pixel by pixel to simulate a survival risk by $h(I) = exp\left(-\sum_{1 \le i \le W, 1 \le j \le H}(I_{ij} * M_{ij})^2\right)$, where the subscript $(i,j)$ indicates the pixel location, as shown in Fig. 2a. Survival time was generated as $\lambda_0 exp(-h(I))$, where $\lambda_0 = 5$ for this simulated dataset. Therefore, the simulated images had survival risks nonlinearly correlated with the image data. Moreover, 50% of the samples were randomly selected as non-censored samples.

We applied the proposed model, the CPH model, and the RSF model to the simulated dataset under the same 10-fold cross-validation setting, our method obtained a c-index of 0.96 while the CPH model obtained a c-index of 0.86 and the RSF model obtained a c-index of 0.85, indicating that our method could achieve improved performance compared with survival models built upon hand-crafted imaging features.

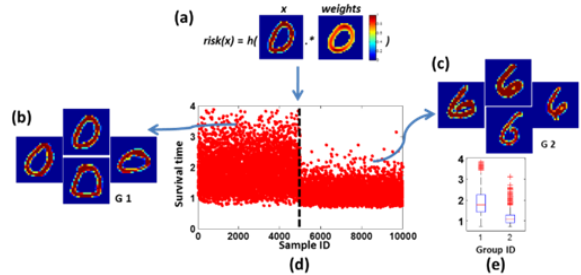

Fig. 2. Simulated images with tumors whose survival risk was a nonlinear function of image intensity information. (a) tumor weights used to generate risks, (b, c) randomly generated images, (d) simulated survival time of those images, and (e) their group difference in survival time.

### 3.2. Experiment on a rectal cancer dataset

We further evaluated our method based on a dataset of 84 rectal cancer patients who received CRT for locally advanced rectal cancer (33 females and 51 males), for predicting time of local tumor recurrence based on their pre-treatment imaging data. All the patients underwent pre-treatment FDG PET/CT scans, and 14 patients had tumor recurrence over a median follow-up of 3 years. The tumors were manually contoured by experienced radiologists. The PET/CT scans of randomly selected patients with/without tumor recurrence are illustrated in Fig. 3. The CT scans were aligned to its corresponding PET scans and resampled to have the same spatial resolution as the PET scans with isotropic voxel size of 4mm. Standardized uptake value (SUV) was computed for the PET scans and used as input to all the survival prediction models.

To augment the training data, augmented tumor images were generated based on the training data by shifting the center of the bounding box by two voxels in 3D space (26 directions), and the tumor recurrence time for the augmented data was sampled according to a Gaussian distribution with the tumor recurrence time of the original tumor $si_o$ as mean value and $0.05si_o$ as standard deviation.

All the survival prediction models were validated using the same 5-fold cross-validation. We tested the prediction performance of different models using PET, CT, and PET/CT scans respectively, and c-index values under different settings are summarized in Table 1. The models built upon multimodal imaging data had better performance than those built upon PET or CT scan alone, and our method achieved overall better prediction performance than the CPH and RSF models. The prediction performance is promising, taking into consideration of the relatively small number of non-censored data in our current dataset.

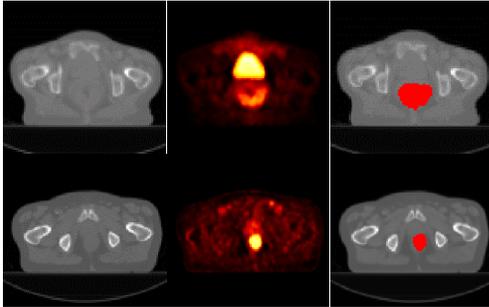

Fig. 3. Images of sample patients with (top row) and without (bottom row) tumor recurrence (CT, PET, and manual tumor segmentation overlaid on the CT image, from left to right).

Table 1. Performance (c-index) of different models using imaging data of different modalities.

|  | CT | PET | PET/CT |
|---|---|---|---|
| CPH | 0.53 | 0.58 | 0.60 |
| RSF | 0.58 | 0.61 | 0.58 |
| proposed | 0.62 | 0.60 | 0.64 |

## 4. CONCLUSIONS

In this study, we develop a flexible deep learning framework to learn reproducible and discriminative radiomic features in a proportional hazards model setting. Our deep learning model is built upon CNNs to learn informative features from tumor images by optimizing partial likelihood of a proportional hazards model. The prediction performance achieved by our method on both the simulated imaging data and the real PET/CT scans of rectal cancer patients is promising, compared to state-of-the-arts methods. The flexibility of our framework also leaves room to fine tune hyper-parameters of the deep learning model and incorporate non-imaging data in the model.

## 5. ACKNOWLEDGEMENTS

This study was supported in part by National Institutes of Health grants [CA223358, EB022573].

## 6. REFERENCES

[1] M. Maas *et al.*, "Long-term outcome in patients with a pathological complete response after chemoradiation for rectal cancer: a pooled analysis of individual patient data," *Lancet Oncology,* vol. 11, no. 9, pp. 835-844, Sep 2010.
[2] I. Joye *et al.*, "Can clinical factors be used as a selection tool for an organ-preserving strategy in rectal cancer?," *Acta Oncologica,* vol. 55, no. 8, pp. 1047-1052, 2016/08/02 2016.
[3] K. Nie *et al.*, "Rectal Cancer: Assessment of Neoadjuvant Chemoradiation Outcome based on Radiomics of Multiparametric MRI," *Clinical Cancer Research,* vol. 22, no. 21, pp. 5256-5264, Nov 1 2016.
[4] N. Dinapoli *et al.*, "Radiomics for rectal cancer," *Translational Cancer Research,* vol. 5, no. 4, pp. 424-431, 2016.
[5] P. Lambin *et al.*, "Radiomics: extracting more information from medical images using advanced feature analysis," *Eur J Cancer,* vol. 48, no. 4, pp. 441-6, Mar 2012.
[6] R. J. Gillies, P. E. Kinahan, and H. Hricak, "Radiomics: Images Are More than Pictures, They Are Data," *Radiology,* vol. 278, no. 2, pp. 563-577, 2016.
[7] M. Vallieres, C. R. Freeman, S. R. Skamene, and I. El Naqa, "A radiomics model from joint FDG-PET and MRI texture features for the prediction of lung metastases in soft-tissue sarcomas of the extremities," *Phys Med Biol,* vol. 60, no. 14, pp. 5471-96, Jul 21 2015.
[8] D. Nie, H. Zhang, E. Adeli, L. Liu, and D. Shen, "3D Deep Learning for Multi-modal Imaging-Guided Survival Time Prediction of Brain Tumor Patients," *Med Image Comput Comput Assist Interv,* vol. 9901, pp. 212-220, Oct 2016.
[9] V. Gulshan, L. Peng, M. Coram, and et al., "Development and validation of a deep learning algorithm for detection of diabetic retinopathy in retinal fundus photographs," *JAMA,* vol. 316, no. 22, pp. 2402-2410, 2016.
[10] W. Shen *et al.*, "Learning from Experts: Developing Transferable Deep Features for Patient-Level Lung Cancer Prediction," in *Medical Image Computing and Computer-Assisted Intervention – MICCAI 2016: 19th International Conference, Athens, Greece, October 17-21, 2016, Proceedings, Part II*, S. Ourselin, L. Joskowicz, M. R. Sabuncu, G. Unal, and W. Wells, Eds. Cham: Springer International Publishing, 2016, pp. 124-131.
[11] I. Goodfellow, Y. Bengio, and A. Courville, *Deep Learning*. MIT Press, 2016.
[12] K. M. He, X. Y. Zhang, S. Q. Ren, and J. Sun, "Spatial Pyramid Pooling in Deep Convolutional Networks for Visual Recognition," *Ieee Transactions on Pattern Analysis and Machine Intelligence,* vol. 37, no. 9, pp. 1904-1916, Sep 2015.
[13] D. R. Cox, "Regression Models and Life-Tables," *Journal of the Royal Statistical Society Series B-Statistical Methodology,* vol. 34, no. 2, pp. 187-+, 1972.
[14] H. Ishwaran, U. B. Kogalur, E. H. Blackstone, and M. S. Lauer, "Random Survival Forests," *Annals of Applied Statistics,* vol. 2, no. 3, pp. 841-860, Sep 2008.
[15] R. Ranganath, A. Perotte, N. Elhadad, and D. Blei, "Deep Survival Analysis," *arXiv:1608.02158,* pp. 1-13, 2016.
[16] J. Katzman, U. Shaham, J. Bates, A. Cloninger, T. Jiang, and Y. Kluger, "Deep Survival: A Deep Cox Proportional Hazards Network," *arXiv:1606.00931v2* 2016.
[17] X. Zhu, J. Yao, and J. Huang, "Deep convolutional neural network for survival analysis with pathological images," in *Bioinformatics and Biomedicine (BIBM), 2016 IEEE International Conference on*, 2016, pp. 544-547: IEEE.
[18] F. Chollet. (2015). *Keras*. Available: https://github.com/fchollet/keras
[19] F. E. Harrell, K. L. Lee, R. M. Califf, D. B. Pryor, and R. A. Rosati, "Regression modelling strategies for improved prognostic prediction," *Statistics in medicine,* vol. 3, no. 2, pp. 143-152, 1984.